\documentclass{styles/svproc}

\usepackage{type1cm}
\usepackage{makeidx}
\usepackage{graphicx}
\usepackage{multicol}
\usepackage[bottom]{footmisc}
\usepackage{siunitx}
\usepackage{caption}
\usepackage{subcaption}
\usepackage{pgfplots}
\pgfplotsset{compat=1.18}
\usepackage{tikz}
\usepgfplotslibrary{units}

\usepackage{newtxtext}
\usepackage[varvw]{newtxmath}

\usepackage{psfrag}
\usepackage{color}
\usepackage[T1]{fontenc}

\makeindex

\usepackage{cite}
\usepackage{url}

\usepackage{comment}
\usepackage{subcaption}

\begin{document}
\mainmatter              

\title{Dynamic Modeling of a Welding Torch Umbilical and Its Impact on Robot Dynamics}
\titlerunning{Dynamic Modeling of a Welding Torch Umbilical and Its Impact on Robot Dynamics}


\author{Nicolas Gautier${}^{1,2}$ , Yves Guillermit${}^{1}$, Mathieu Porez${}^{2}$,\\ Fabien Rousset${}^{1}$ and Damien Chablat${}^{2}$}

\authorrunning{Nicolas Gautier, Yves Guillermit, Mathieu Porez, Fabien Rousset and Damien Chablat}

\institute{Weez-U Welding, 41-43 Quai de Malakoff, 44 000 Nantes, France  \{\email{ngr@weez-u-welding.com}, \email{ygt@weez-u-welding.com}, \email{frt@weez-u-welding.com}\} \and  Nantes Université, École Centrale Nantes, Institut Mines Télécom Atlantique, CNRS, LS2N, UMR 6004, F-44000 Nantes, France \{\email{mathieu.porez@imt-atlantique.fr}, \email{damien.chablat@cnrs.fr}\}}

\maketitle

\begin{abstract}
Robotic welding is widely used in industrial manufacturing, where the welding torch is often connected to the generator through an external umbilical. With the increasing deployment of lightweight and collaborative robots, the dynamic influence of this umbilical can significantly affect the robot motion and the actuation forces. This paper proposes a constrained multibody dynamic model of a welding umbilical, represented as a serial chain of rigid bodies interconnected by passive joints with elastic and dissipative effects. Prescribed motions at the distal anchor point are introduced through holonomic kinematic constraints. The equations of motion are reduced by projecting the dynamics onto the subspace of admissible velocities, yielding an efficient formulation free of Lagrange multipliers. The reaction wrench exerted by the umbilical on the robot is explicitly recovered. A planar case study illustrates the approach.
\keywords{Robotic Welding, Multibody dynamics, Constrained systems, Flexible cable modeling}
\end{abstract}
\section{Introduction}
Robotic welding is widely used in industrial manufacturing due to its ability to provide high productivity, repeatability, and consistent weld quality. In such applications, the welding torch mounted on the robot end-effector is connected to the welding generator through a welding umbilical \cite{weman_welding_2011, pires_welding_2006}. Depending on the robot design and workspace constraints, the welding umbilical may be routed internally or externally along the robot structure. While external routing remains common in industrial setups, it can be problematic for lightweight and collaborative robots. Due to their reduced mass, these robots are more sensitive to external loads. In such context, the welding umbilical can no longer be regarded as a negligible component. Its weight and stiffness (torsion and bending) introduce additional forces that significantly affect the robot's dynamics. Accurately modeling the welding umbilical is therefore essential for applications requiring reliable force estimation and realistic dynamic response, such as control, collision detection, and physical human–robot interaction, provided that the resulting model remains computationally efficient and compatible with in-line integration \cite{khalil2004modeling}.

The dynamic modeling of elongated flexible structures, such as welding umbilicals, has been extensively studied in the literature using a variety of approaches. Continuous formulations—based on finite element methods or geometrically exact beam theories—provide high-fidelity representations of cable dynamics \cite{tian2019modelling,quan2013three}. However, despite their accuracy, these methods often result in large-scale systems that are computationally expensive. Another well-established framework relies on Cosserat rod theory, which models cables and slender structures as continuous rods with distributed strain variables \cite{khalil2007dynamic, boyer2006macro,boyer2023implicit}.
While highly expressive, these models often require implicit integration schemes and careful numerical treatment, especially to solve the forward problem of dynamics. To reduce computational complexity, discretized multibody or lumped-parameter models can be used. The cable is approximated as a serial chain of rigid links connected by passive joints with stiffness and damping \cite{klimchik2025stiffness, du2022dynamic, moussa2025dynamic}. This strategy provides a favorable trade-off between physical realism and computational efficiency. More recently, hybrid or data-driven approaches have been explored. In \cite{mou2022learning}, the authors combine physics-based modeling with learning techniques to capture complex coupling effects between the robot and heavy industrial cables.
Motivated by the need for a computationally efficient and physically meaningful model of welding cable, this work adopts a multibody discretization of the umbilical as a serial chain of rigid bodies interconnected by passive joints. Elasticity and dissipation are introduced through joint stiffness and friction models, while prescribed kinematic constraints are imposed at the distal anchor point. The resulting constrained dynamics are formulated using Lagrange multipliers and subsequently reduced by projecting the equations of motion onto the subspace of admissible velocities. This projection-based approach eliminates the constraint forces from the reduced system, enabling an efficient computation of joint accelerations, while preserving the ability to recover the reaction wrench at the anchor point.

The remainder of the paper is organized as follows. Sections 2 \& 3 present the general kinematic and dynamic formulation of the umbilical as a constrained multibody system. Section 4 illustrates the approach on a planar case study. Finally, conclusions and perspectives are discussed in Section 5.
\section{Kinematics Model of the umbilical}
\begin{figure}[!ht]
    \centering
    \includegraphics[width=0.8\linewidth]{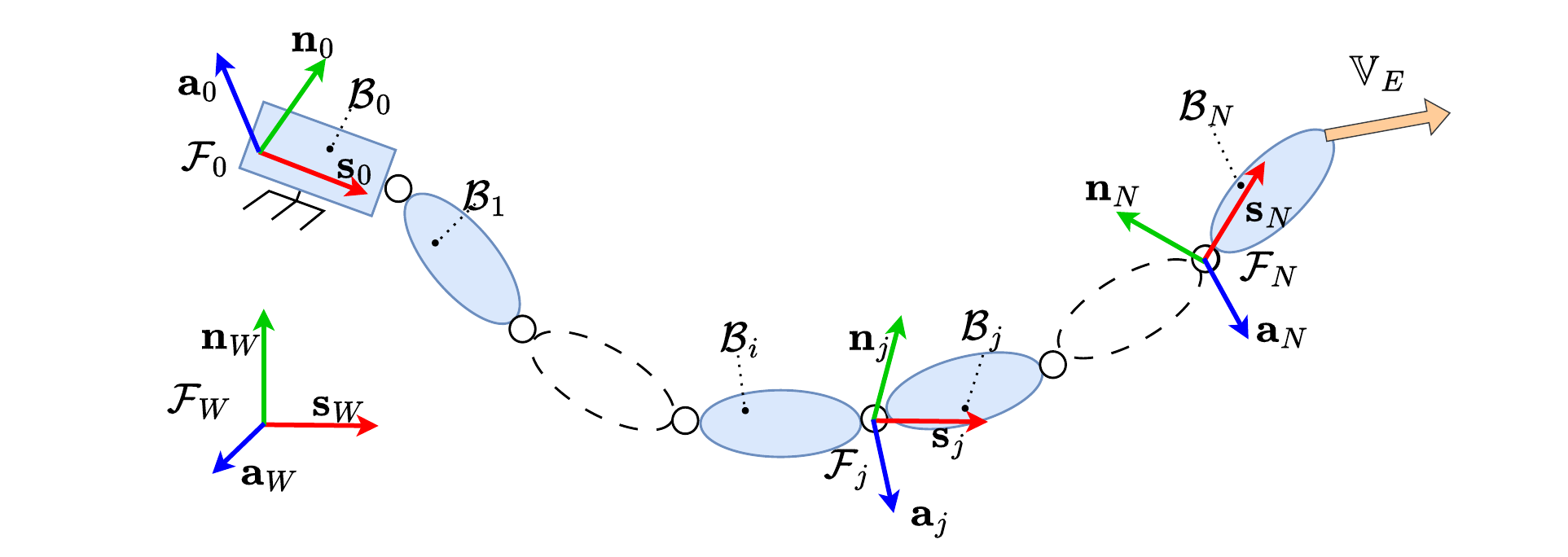}
    \vspace{-0.3cm}
    \caption{Definition of world, umbilical, and anchor point frames. The TCP frame follows the prescribed motions.}
    \label{fig:bodies}
\end{figure}

The umbilical is modeled as a serial multibody system composed of $N+1$ rigid bodies interconnected by $N$ joints. Let us remark that the type of joints and their assembly depend of the beam kinematic that we want to reproduce. The bodies are indexed from the fixed base body $\mathcal{B}_0$ to the distal body $\mathcal{B}_N$. For a given body $\mathcal{B}_j$, the index $i$ denotes its antecedent body. An orthonormal frame $\mathcal{F}_j = (O_j,\mathbf{s}_j,\mathbf{n}_j,\mathbf{a}_j)$ is attached to each body, where the unit vector $\mathbf{a}_j$ defines the joint axis. The configuration of the system is described by the vector of generalized coordinates $\mathbf{q} = (q_1,\ldots,q_N)^T \in \mathbb{R}^N$, where $q_j$ denotes the relative angles around the joint axis $j$.
The two tips of the umbilical are assumed to be clamped. The base body $\mathcal{B}_0$ is fixed with respect to the world frame $\mathcal{F}_w$, while the distal body $\mathcal{B}_N$ may be subjected to prescribed motions (e.g.when attached to a robotic manipulator). This induces a set of $k$ holonomic constraints at the configuration level. For numerical implementation, these constraints are enforced at the velocity level by enforcing a prescribed velocity twist $\mathbb{V}_E \in \mathbb{R}^{6}$. The constraint function $\boldsymbol{\psi}(\mathbf{q}, \dot{\mathbf{q}}, t)$ is defined as:
\begin{equation}
 \boldsymbol{\psi}(\mathbf{q},\dot{\mathbf{q}},t) = \mathbf{A}(\mathbf{q})\dot{\mathbf{q}} - \mathbb{V}_E = \mathbf{0} \text{ ,}
\label{eq:constraints}
\end{equation}
where $\mathbf{A}(\mathbf{q})$ corresponds to the kinematic Jacobian matrix with respect to the generalized coordinates~\cite{khalil2004modeling}.  The time derivative $\dot{\mathbf{A}}(\mathbf{q},\dot{\mathbf{q}})$ naturally arises in the acceleration-level formulation and will be used in the dynamic model. The complete system is illustrated in Fig.~\ref{fig:bodies}.
\vspace{-0.5cm}
\vspace{-1cm}
\section{Dynamic Model of the Umbilical}
We consider the following forward dynamics problem. At each time $t$, the state of the multibody system is assumed to be known and is defined by $\mathbf{x}^T = (\dot{\mathbf{q}}^T,\mathbf{q}^T)$. The objective is to compute the joint accelerations $\ddot{\mathbf{q}}$ by solving the forward dynamic equations of the constrained multibody system together with the wrench of the reaction forces $\mathbf{f}_N$ applied at the distal anchor point. The equations of motion are expressed in an assembled Lagrangian form as \cite{6710155}:

\begin{equation}
\mathbf{M}_0(\mathbf{q}) \ddot{\mathbf{q}}
+ \mathbf{Q}_0(\mathbf{q}, \dot{\mathbf{q}})
- \boldsymbol{\tau}
= \nabla_{\dot{\mathbf{q}}} \boldsymbol{\psi}(\mathbf{q}, \dot{\mathbf{q}}, t)^T \boldsymbol{\lambda} \text{ ,}
\label{eq:forward_dynamics}
\end{equation}

\noindent  where $\mathbf{M}_0 \in\mathbb{R}^{N\times N}$ denotes the inertia matrix of the system, $\mathbf{Q}_0 \in\mathbb{R}^{N}$ gathers the vector of the generalized forces (including Coriolis, centrifugal, and gravitational effects), and $\boldsymbol{\tau} \in\mathbb{R}^{N}$ represents the vector of joint torques arising from the elastic stiffness and damping properties of the umbilical.  The vector $\boldsymbol{\lambda}\in\mathbb{R}^{k}$ gathers the associated Lagrange multipliers and represents the generalized constraint forces. These forces correspond to the reaction wrench transmitted from the anchor point to the multibody system through the persistent contact constraints.

The joints of the umbilical are passive and not directly actuated. The joint torques $\boldsymbol{\tau}$ therefore arise solely from the mechanical properties of the umbilical, namely its elastic stiffness and internal friction effects. The resulting joint torque vector is modeled as:

\begin{equation}
\boldsymbol{\tau} = - \mathbf{K} \mathbf{q} - \mathbf{F}_c \mathrm{sign}(\dot{\mathbf{q}}) -\mathbf{F}_v  \dot{\mathbf{q}} \text{ ,}
\label{eq:joint_torque}
\end{equation}

\noindent  where $\mathbf{K} \in \mathbb{R}^{N\times N}$ is the joint stiffness diagonal matrix, and $\mathbf{F}_c,\mathbf{F}_v \in \mathbb{R}^{N\times N}$ are diagonal matrices gathering the Coulomb and viscous friction coefficients, respectively.

To compute the various dynamic matrices involved in the formulation~\eqref{eq:forward_dynamics}, the Newton–Euler formalism is employed. This approach has been extensively used for both forward dynamics~\cite{Featherstone} and inverse dynamics~\cite{luh1980line} of multibody systems, owing to its recursive structure, which ensures computational efficiency and scalability with respect to the number of degrees of freedom. 
The Newton–Euler formulation allows the systematic computation of link velocities, accelerations, and internal forces through forward and backward recursive calculations along the kinematic chain. In the present work, it is used to evaluate the inertia matrix $\mathbf{M}_0$, and the vector of generalized internal forces $\mathbf{Q}_0$. The detailed derivation of the recursive equations and their implementation can be found in~\cite{mauny2017symbolic}. For the sake of conciseness, only the resulting assembled equations are presented in this paper.
In order to reduce the dynamics of the constrained system, ~\eqref{eq:forward_dynamics} is projected onto the null space of ~\eqref{eq:constraints}, corresponding to the subspace of admissible joint velocities. To this end, the generalized velocity vector is partitioned into dependent (constrained) and independent components, $\dot{\mathbf{q}}^T = (\dot{\mathbf{q}}_d^T, \dot{\mathbf{q}}_i^T)$, where the subscripts $d$ and $i$ denote the dependent and independent joints, respectively. The independent coordinates are those dynamically solved in the reduced system, while the dependent coordinates are obtained from the constraint equations. Accordingly, the constraint Jacobian is decomposed as $\mathbf{A} = \begin{bmatrix} \mathbf{A}_d & \mathbf{A}_i  \end{bmatrix}$, with $\mathbf{A}_d \in \mathbb{R}^{k\times k}$ and $\mathbf{A}_i \in \mathbb{R}^{k\times (N-k)}$ \cite{merlet2006parallel}. Assuming that $\mathbf{A}_d$ is nonsingular, the velocity constraint~\eqref{eq:constraints} can be solved for the dependent joint velocities as:
\begin{equation}
 \dot{\mathbf{q}}_d    = \begin{bmatrix} -\mathbf{A}_d^{-1}\mathbf{A}_i \quad & \mathbf{A}_d^{-1}  \end{bmatrix} \begin{pmatrix} \dot{\mathbf{q}}_i \\ \mathbb{V}_E   \end{pmatrix} \text{ .}
\end{equation}

Thus, the admissible joint velocity vector takes the compact form:
\begin{equation}
\dot{\mathbf{q}} = \mathbf{H} \begin{pmatrix} \dot{\mathbf{q}}_i \\ \mathbb{V}_E \end{pmatrix}\text{, and }\mathbf{H} \in\mathbb{R}^{N\times N} =\begin{bmatrix} -{\mathbf{A}_d}^{-1}\mathbf{A}_i \quad & {\mathbf{A}_d}^{-1} \\\mathbf{1} & \mathbf{0}   \end{bmatrix} \text{,}
\label{eq:admissible_velocity}\end{equation}
where $\mathbf{1}$ and $\mathbf{0}$ denotes respectively the identity matrix and the null matrix. Differentiating~\eqref{eq:admissible_velocity} with respect to time yields the joint acceleration vector:
\begin{equation}
\ddot{\mathbf{q}} = \dot{\mathbf{H}} \begin{pmatrix} \dot{\mathbf{q}}_i \\ \mathbb{V}_E \end{pmatrix}+ \mathbf{H} \begin{pmatrix} \ddot{\mathbf{q}}_i \\ \dot{\mathbb{V}}_E \end{pmatrix} \text{,}\label{eq:admissible_acceleration}
\end{equation}
where $\dot{\mathbf{H}}$ denotes the time derivative of the admissible velocity mapping $\mathbf{H}$. Using the definition of $\mathbf{H}$ given in~\eqref{eq:admissible_velocity}, the expression of $\dot{\mathbf{H}}$ can be analytically obtained as:
\begin{equation}
\dot{\mathbf{H}}  = \begin{bmatrix} {\mathbf{A}_d}^{-1}(\dot{\mathbf{A}}_d {\mathbf{A}_d}^{-1}\mathbf{A}_i- \dot{\mathbf{A}}_i) \quad& -{\mathbf{A}_d}^{-1}\dot{\mathbf{A}}_d {\mathbf{A}_d}^{-1} \\ \mathbf{0} & \mathbf{0}   \end{bmatrix} \text{ .}
\label{eq:Hdot}
\end{equation}

Substituting~\eqref{eq:admissible_acceleration} into~\eqref{eq:forward_dynamics}, and projecting the resulting equations onto the subspace of admissible velocities using the matrix $\mathbf{H}$, leads to a reduced formulation in which the Lagrange multipliers do not appear explicitly. By construction, the first $N-k$ columns of $\mathbf{H}$ span the null space of $\mathbf{A}(\mathbf{q})$, while the remaining $k$ columns ensure that the velocity constraint $\mathbf{A}(\mathbf{q})=\mathbb{V}_E$ is satisfied. Consequently, premultiplying the equations by $\mathbf{H}^T$ eliminates the constraint forces associated with the admissible velocity subspace. The resulting reduced dynamic system is given by:
\begin{equation}
\mathbf{M}_r \begin{pmatrix} \ddot{\mathbf{q}}_i \\ \dot{\mathbb{V}}_E \end{pmatrix} + \mathbf{Q}_r = \mathbf{0} \text{,}
\label{eq:reduced_dynamic}
\end{equation}
where the reduced inertia matrix $\mathbf{M}_r \in \mathbb{R}^{N \times N}$ and the reduced force vector $\mathbf{Q}_r \in \mathbb{R}^{N}$ are defined as:

\begin{equation}
\mathbf{M}_r = \mathbf{H}^T \mathbf{M}_0 \mathbf{H} \text{ , and }
\mathbf{Q}_r = \mathbf{H}^T \left( \mathbf{Q}_0 - \boldsymbol{\tau} + \mathbf{M}_0 \dot{\mathbf{H}} \begin{pmatrix} \dot{\mathbf{q}}_i \\ \mathbb{V}_E \end{pmatrix}
\right) \text{.}
\end{equation}

In~\eqref{eq:reduced_dynamic}, the term $\dot{\mathbb{V}}_E$ is known, as it corresponds to the prescribed vector of acceleration imposed at the distal anchor point. The unknown quantity to be determined is therefore the vector of independent joint accelerations $\ddot{\mathbf{{q}}}_i$. The reduced matrices and vectors are partitioned accordingly as:
\begin{equation}
\mathbf{M}r = \begin{bmatrix} \mathbf{M}_{\alpha_1} & \mathbf{M}_{\beta_1} \\ \mathbf{M}_{\alpha_2} & \mathbf{M}_{\beta_2} \end{bmatrix}\text{, and }
\mathbf{Q}_r = \begin{pmatrix} \mathbf{Q}_1 \\ \mathbf{Q}_2 \end{pmatrix},
\label{eq:reduced_dynamic_decompose}
\end{equation}
leading to the linear system:
\begin{equation}
\mathbf{M}_{\alpha_1}\ddot{\mathbf{q}}_i = -\mathbf{Q}_1 -\mathbf{M}_{\beta_1}\dot{\mathbb{V}}_E.
\label{eq:independent_acceleration}
\end{equation}

The linear system is solved to obtain the independent joint accelerations $\ddot{\mathbf{q}}_i$. The full joint acceleration vector $\ddot{\mathbf{q}}$ is then reconstructed using~\eqref{eq:admissible_acceleration}. Finally, the reaction forces exerted by the umbilical at the anchor point can be recovered from the Lagrange multipliers. According to~\eqref{eq:forward_dynamics}, the multipliers satisfy:
\begin{equation}
\boldsymbol{\lambda} = {(\mathbf{A}^T(\mathbf{q}))}^+(\mathbf{M}_0(\mathbf{q})\ddot{\mathbf{q}} + \mathbf{Q}_0(\mathbf{q}, \dot{\mathbf{q}})-\boldsymbol{\tau}) \text{ .}
\label{eq:lambda_equation}
\end{equation}
where $(\cdot)^+$ denotes the Moore–Penrose pseudo-inverse. This formulation provides the minimum-norm solution of the Lagrange multipliers under the assumption that the constraint Jacobian $\mathbf{A}(\mathbf{q})$ has full row rank. The resulting vector $\boldsymbol{\lambda}$ corresponds to the generalized reaction wrench transmitted by the umbilical at the anchor point. These reaction forces can therefore be directly used as input for the dynamic model of the robot to which the umbilical is attached, ensuring a consistent coupling between the umbilical dynamics and the robot motion.
\section{Application to a Planar Case Study}
To illustrate the proposed formulation and assess the dynamic impact of a welding umbilical on a robotic system, we consider a planar case study where closed-form expressions for the kinematic quantities can be derived. The umbilical of length $L$ is modeled as a planar serial chain consisting of $N$ revolute joints and $N+1$ rigid links of equal length $l=L/(N+1)$. This discretized representation is equivalent to a planar Kirchhoff–Love beam model, where only in-plane bending deformations are considered. Axial extension, compression, and transverse shear effects are neglected, and the links are assumed inextensible. The Jacobian matrix $\mathbf{A} \in \mathbb{R}^{3 \times N}$, which relates the joint velocities to  $\mathbb{V}_E$, is obtained by differentiating the forward geometric model~\cite{khalil2004modeling}. The $i$-th column of the Jacobian is given by:

\begin{equation}
    {}^0\mathbf{A}_{i,N} = \begin{bmatrix} -l\sum_{j=i}^N {\sin({\phi}}_j) & \quad l\sum_{j=i}^N {\cos({\phi}}_j) & \quad1\end{bmatrix}^T \text{ with } \phi_j = \sum_{k=1}^j q_k \text{.}
    \label{eq:jacobian}
\end{equation}

The time derivative of the Jacobian, required for the constrained dynamic formulation, is obtained by direct differentiation. The $i$-th column of $\dot{\mathbf{A}}$ is:
\begin{equation}
    {}^0\dot{\mathbf{A}}_{i,N} = \begin{bmatrix} -l\sum_{j=i}^N {\cos({\phi}}_j) \dot{\phi}_j \quad & -l\sum_{j=i}^N {\sin({\phi}}_j) \dot{\phi}_j \quad& 0\end{bmatrix}^T \text{ .}
    \label{eq:jacobian_dot}
\end{equation}

\begin{figure}

	\centering
    \hspace*{-0.4cm}
	\begin{subfigure}{0.32\textwidth}
        \begin{tikzpicture}
\hspace*{-0.3cm}
\begin{axis}[
    axis equal image,
    width=6cm,
    xmin=-0.2, xmax=1.2,
    ymin=-0.6, ymax=0.8,
    axis lines=box,                
    xtick distance=0.4,            
    ytick distance=0.4,            
    xticklabels={},                 
    yticklabels={},
    grid=major,                     
    major grid style={line width=0.8pt, draw=gray!70},
    minor tick num=0,                
    legend style={font=\scriptsize, at={(0.8,1.1)}, anchor=north east}
]
\addplot[
    red,
    dashed,
    line width=2pt,
    domain=0:360,
    samples=200
] (
    {0.7 + 0.3*cos(x)},
    {0.00 + 0.3*sin(x)}
);
\addlegendentry{Trajectory}
\draw[fill = gray] (-0.2,-0.2) rectangle (0,0.4);
\addplot[
    blue,
    line width=2pt,
    mark=*,
    mark size=1.5pt
] coordinates {
    (0.0,0.0)
    (0.14,0.0)
    (0.23493,-0.1029)
    (0.28056,-0.23525)
    (0.32903,-0.36659)
    (0.42907,-0.46453)
    (0.56906,-0.46619)
    (0.67111,-0.37035)
    (0.72151,-0.23973)
    (0.76827,-0.10777)
    (0.86356,-0.00521)
    (1.00356,-0.00521)
};
\addlegendentry{Umbilical}

\addplot[
    black,
    line width=1.2pt,
    double,
    double distance=1.2pt,
    mark=o,
    mark size=5pt,
    mark options={
        draw=black,
        fill=white,
        line width=1pt
    },
    mark layer=axis foreground
] coordinates {
    (0.0, 0.2)
    (0.35663,  0.38115)
    ( 0.74865,  0.30169)
    (1.00351, -0.00661)
};
\addlegendentry{Robot}

\end{axis}
\end{tikzpicture} 
        {\vspace{-0.5cm}}
		\caption{Initial configuration and circular end-effector trajectory.}
	\end{subfigure}
   \hfill
   \hspace*{-0.6cm}
	\begin{subfigure}{0.3\textwidth}
        \center
        \begin{tikzpicture}
\hspace*{-0.2cm}
\begin{axis}[
    axis equal image,
    width=3.92cm,
    xmin=-0.25, xmax=0.9,
    ymin=-0.7, ymax=0.7,
    axis lines=box,
    ticks=none,
    clip=false
]
\addplot[
    blue,
    line width=1pt,
    mark=*,
    mark size=0.8pt
] coordinates {
(0,0)
(0.14   ,  0.     )
(0.23157, -0.1059 )
(0.29178, -0.23229)
(0.3816 , -0.33968)
(0.52013, -0.35995)
(0.61717, -0.25904)
(0.62623, -0.11933)
(0.59759,  0.01771)
(0.5884 ,  0.15741)
(0.65587,  0.28008)
(0.79587,  0.28008)
};

\addplot[
    black,
    line width=0.6pt,
    double,
    double distance=0.6pt,
    mark=o,
    mark size=2.5pt,
    mark options={
        draw=black,
        fill=white,
        line width=0.6pt
    },
    mark layer=axis foreground
] coordinates {
    (0.    ,  0.2    )
(0.14143, 0.57416)
(0.54122, 0.58708)
(0.79608, 0.27879)
};

\addplot[
    red,
    dashed,
    line width=1pt,
    domain=70:300,
    samples=200
] (
    {0.7 + 0.3*cos(x)},
    {0.00 + 0.3*sin(x)}
);
\end{axis}

\begin{axis}[
    at={(1.54cm,0cm)},
    axis equal image,
    width=3.92cm,
    xmin=-0.25, xmax=0.9,
    ymin=-0.7, ymax=0.7,
    axis lines=box,
    ticks=none,
    clip=false
]
\addplot[
    red,
    dashed,
    line width=1pt,
    domain=70:300,
    samples=200
] (
    {0.7 + 0.3*cos(x)},
    {0.00 + 0.3*sin(x)}
);
\addplot[
    blue,
    line width=1pt,
    mark=*,
    mark size=0.8pt
] coordinates {
(0,0)
    (0.14   ,  0.     )
(0.20261, -0.12522)
(0.2142 , -0.26474)
(0.25689, -0.39807)
(0.38672, -0.45045)
(0.47021, -0.33807)
(0.42943, -0.20414)
(0.34726, -0.09079)
(0.28594,  0.03507)
(0.32027,  0.17079)
(0.46027,  0.17079)
};
\addplot[
    black,
    line width=0.6pt,
    double,
    double distance=0.6pt,
    mark=o,
    mark size=2.5pt,
    mark options={
        draw=black,
        fill=white,
        line width=0.6pt
    },
    mark layer=axis foreground
] coordinates {
    ( 0.     ,  0.2    )
(-0.18693,  0.55364)
( 0.20584,  0.47793)
( 0.4607 ,  0.16963)
};

\end{axis}

\begin{axis}[
    at={(0cm,-1.86cm)},
    axis equal image,
    width=3.92cm,
    xmin=-0.25, xmax=0.9,
    ymin=-0.7, ymax=0.7,
    axis lines=box,
    ticks=none,
    clip=false
]
\addplot[
    red,
    dashed,
    line width=1pt,
    domain=70:300,
    samples=200
] (
    {0.7 + 0.3*cos(x)},
    {0.00 + 0.3*sin(x)}
);
\addplot[
    blue,
    line width=1pt,
    mark=*,
    mark size=0.8pt
] coordinates {
(0,0)
    (0.14   ,  0.     )
(0.17408, -0.13579)
(0.11181, -0.26118)
(0.02753, -0.37297)
(0.01803, -0.50535)
(0.05934, -0.62203)
(0.19176, -0.57661)
(0.23975, -0.44509)
(0.25505, -0.30593)
(0.31976, -0.18178)
(0.45976, -0.18178)
};
\addplot[
    black,
    line width=0.6pt,
    double,
    double distance=0.6pt,
    mark=o,
    mark size=2.5pt,
    mark options={
        draw=black,
        fill=white,
        line width=0.6pt
    },
    mark layer=axis foreground
] coordinates {
    (0.     ,  0.2    )
(0.23427,  0.52422)
(0.20581,  0.12524)
(0.46067, -0.18306)
};

\end{axis}

\begin{axis}[
    at={(1.54cm,-1.86cm)},
    axis equal image,
    width=3.92cm,
    xmin=-0.25, xmax=0.9,
    ymin=-0.7, ymax=0.7,
    axis lines=box,
    ticks=none,
    clip=false
]
\addplot[
    red,
    dashed,
    line width=1pt,
    domain=70:300,
    samples=200
] (
    {0.7 + 0.3*cos(x)},
    {0.00 + 0.3*sin(x)}
);
\addplot[
    blue,
    line width=1pt,
    mark=*,
    mark size=0.8pt
] coordinates {
(0,0)
    (0.14   ,  0.     )
(0.20811, -0.12232)
(0.19808, -0.26196)
(0.16725, -0.39852)
(0.17313, -0.5384 )
(0.26739, -0.64191)
(0.40639, -0.62517)
(0.49859, -0.51982)
(0.56157, -0.39478)
(0.65502, -0.29054)
(0.79502, -0.29054)
};
\addplot[
    black,
    line width=0.6pt,
    double,
    double distance=0.6pt,
    mark=o,
    mark size=2.5pt,
    mark options={
        draw=black,
        fill=white,
        line width=0.6pt
    },
    mark layer=axis foreground
] coordinates {
    ( 0.     ,  0.2    )
( 0.36054,  0.37325)
( 0.54122,  0.01638)
( 0.79608, -0.29192)
};

\end{axis}

\end{tikzpicture}
		\caption{Intermediate positions of the umbilical.}
	\end{subfigure}
    \hfill
    \hspace*{-0.5cm}
	\begin{subfigure}{0.3\textwidth}       
        \usetikzlibrary{
        matrix,
    }
\begin{tikzpicture}
\hspace*{-0.9cm}

\definecolor{MPLBlue}{RGB}{31,119,180}
\definecolor{MPLOrange}{RGB}{255,127,14}
\definecolor{MPLGreen}{RGB}{44,160,44}
\begin{axis}[
    scale only axis,
    width=4cm,
    xlabel= Time,
    ylabel= Torque,
    x unit=\si{\second},
    y unit=\si{\newton\metre},
    xmin=0.0, xmax=5.0,
    ymin=0.0, ymax=200,
    axis lines=box,                
    xtick distance=1.0,            
    ytick distance=50.0,            
    x label style={at={(axis description cs:0.5,-0.05)}},
    y label style={at={(axis description cs:-0.07,.5)}},
    tick style={major tick length=2pt},
    xticklabel style={inner sep=1pt, font=\scriptsize},
    yticklabel style={inner sep=1pt, font=\scriptsize},
    grid=major,                     
    major grid style={line width=0.8pt, draw=gray!70},
    minor tick num=0,                
    legend style={font=\scriptsize, at={(0.8,1.1)}, anchor=north east}  
]

\addplot[
  solid,
  mark=none,
  color=MPLBlue,
  line width=0.7pt
]
table [x=t, y=tau1, col sep=comma] {img/torque_robot.csv};
\label{plot:line1}

\addplot[
  solid,
  mark=none,
  color=MPLOrange,
  line width=0.7pt
]
table [x=t, y=tau2, col sep=comma] {img/torque_robot.csv};
\label{plot:line2}

\addplot[
  solid,
  mark=none,
  color=MPLGreen,
  line width=0.7pt
]
table [x=t, y=tau3, col sep=comma] {img/torque_robot.csv};
 \label{plot:line3}

\addplot[
  dashed,
  mark=none,
  color=MPLBlue,
  line width=0.7pt
]
table [x=t, y=tau1_without, col sep=comma] {img/torque_robot.csv};
 \label{plot:line4}

\addplot[
  dashed,
  mark=none,
  color=MPLOrange,
  line width=0.7pt
]
table [x=t, y=tau2_without, col sep=comma] {img/torque_robot.csv};
 \label{plot:line5}

\addplot[
  dashed,
  mark=none,
  color=MPLGreen,
  line width=0.7pt
]
table [x=t, y=tau3_without, col sep=comma] {img/torque_robot.csv};
 \label{plot:line6}

\coordinate (legend) at (axis description cs:0.85,0.8);

\end{axis}

\matrix [
    draw,
    fill=white,
    matrix of nodes,
    anchor=south east,
    font=\scriptsize,          
    row sep=1pt,               
    column sep=2pt,            
    inner sep=1pt,             
    nodes={inner sep=0pt},     
] at (legend) {
    Umbilical & No umbilical & \\
    \ref{plot:line1} & \ref{plot:line4} & J1 \\
    \ref{plot:line2} & \ref{plot:line5} & J2 \\
    \ref{plot:line3} & \ref{plot:line6} & J3 \\
};

\end{tikzpicture}
        {\vspace{-0.7cm}}
		\caption{Effect of the welding umbilical on robot joint torques.}
    \end{subfigure}
	\caption{Robot–umbilical interaction in the planar case study.}
	\label{fig:example_planar}
\end{figure}
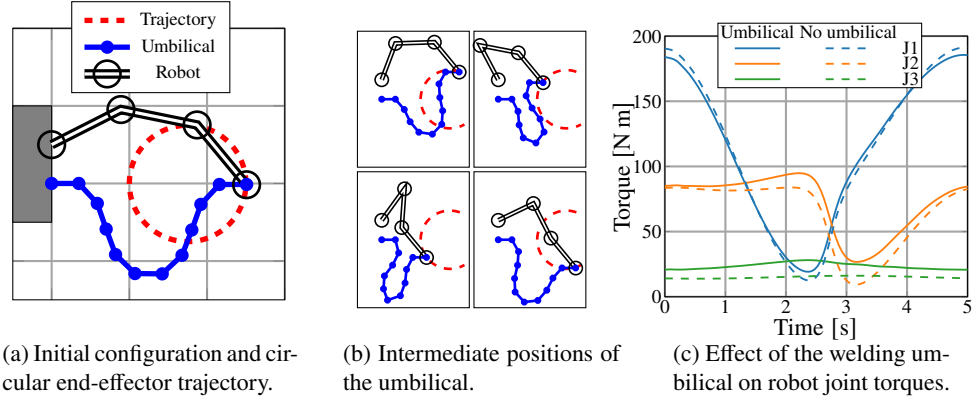

\noindent At the first step of the simulation, the initial configuration of $\mathbf{q}$ must be specified. To minimize artificial transient effects, this configuration is chosen as close as possible to the static equilibrium under gravity. For initialization, the umbilical is modeled as a flexible, homogeneous, and inextensible cable, subjected only to a uniform gravitational field. Under these assumptions, the static equilibrium shape is approximated by a catenary curve~\cite{such2009approach}.
In the current study, the umbilical is discretized into $N=10$ revolute joints with a segment length $l=\qty{0.14}{\m}$, resulting in a total length of \qty{1.54}{\m}. The cable is characterized by a linear mass density of \qty{2}{\kg\per\m}, an equivalent local joint bending stiffness of \qty{5}{\N\m\per\radian} and a joint viscous friction of \qty{0.1}{\N\m\s\per\radian}. 
These values were selected to be consistent with industrial robot welding umbilicals, but no experimental identification was performed at this stage. The choice $N=10$ balances computational cost and deformation resolution. Increasing $N$ did not significantly alter the qualitative results.
The distal end of the umbilical is attached to a planar 3-R robot with equal link lengths of \qty{0.4}{\m} and a total mass of \qty{36}{\kg}. For the needs of the example, the robot dynamics are modeled using an inverse dynamic computed-torque formulation~\cite{luh1980line}. At each simulation step, the robot joint position, velocity, and acceleration profiles are prescribed.
The corresponding end-effector twist and acceleration are obtained through the forward kinematic recursion of the Newton–Euler algorithm. These quantities are then imposed as kinematic constraints in the umbilical model. The constrained direct dynamics of the umbilical are then solved to compute both the joint accelerations and the reaction wrench exerted at the anchor point. This wrench is subsequently introduced in the backward recursion of the robot inverse dynamics to evaluate the additional joint torques induced by the cable. This procedure results in a fully coupled robot–umbilical dynamic simulation.
As a proof of concept, the robot end-effector is required to follow a circular trajectory of radius $R=\qty{0.3}{\m}$ with constant orientation over a duration of \qty{5}{s}. This trajectory is chosen for illustrative purposes and does not restrict the generality of the proposed approach. The desired Cartesian velocity is defined as:
\begin{equation}
\mathbf{V}(t) = \begin{pmatrix} -R\omega \sin(\omega t) \quad& R\omega \cos(\omega t) \quad& 0 \end{pmatrix}^T \text{,}
\end{equation}
which are converted into joint-space commands using the Jacobian pseudoinverse. The linear system~\eqref{eq:independent_acceleration} is solved iteratively using a Krylov subspace method such as GMRES~\cite{saad1986gmres} to avoid the explicit inversion of the matrix $\mathbf{M}_{\alpha_1}$, which may be large and ill-conditioned for a fine discretization of the umbilical. The coupled system is integrated using a predictor–corrector scheme with a fixed time step of \qty{0.5}{\ms}. Figure~\ref{fig:example_planar}(a,b) illustrates the initial and intermediate configurations of the system, including the robot, the welding umbilical, and the prescribed trajectory. Figure~\ref{fig:example_planar}(c) compares the robot joint torques obtained with and without considering the umbilical dynamics. The results show that the torques induced by the umbilical are non-negligible and significantly modify the robot dynamic response, despite the umbilical mass being more than one order of magnitude smaller than that of the robot. These results highlight the importance of explicitly modeling external welding cables in robotic applications, particularly for lightweight robots and dynamic trajectories, as neglecting their influence may lead to significant errors in torque estimation and control performance.
{\vspace{-0.4cm}}
{\vspace{-1cm}}
\section{Conclusions and future work}
This paper introduced a dynamic modeling framework for welding umbilicals and their interaction with robotic manipulators. The umbilical was modeled as a constrained multibody system, consisting of a serial chain of rigid bodies connected by passive joints. This approach captures elastic and dissipative effects while remaining compatible with standard robotic dynamics formalisms. Prescribed motions at the distal anchor point were enforced through holonomic kinematic constraints, and the equations of motion were efficiently reduced by projecting them onto the subspace of admissible velocities. The proposed formulation enables both the computation of the umbilical joint accelerations and the explicit recovery of the reaction wrench exerted on the robot. A planar case study illustrated the approach, demonstrating that the dynamic effects induced by the welding umbilical are non-negligible, even when its mass is significantly lower than that of the robot. These findings underscore the importance of explicitly accounting for external welding cables in robotic control and dynamic analysis, especially for lightweight and collaborative robots.
Future work will focus on experimentally validating the proposed model using a real robotic welding setup and identifying the umbilical's mechanical parameters. Extensions to the framework will also be explored, including attachments at intermediate robot links and the integration of mobile or sliding anchor points along the robot structure.
\section*{Acknowledgment}
This research was supported by ANRT CIFRE grant n°2023 /1565, which funded the first author's doctoral studies. 

\bibliographystyle{ieeetr}
\bibliography{references}
\end{document}